\documentclass[letterpaper, 10 pt, conference]{ieeeconf}  

\IEEEoverridecommandlockouts                              

\usepackage{graphics} 
\usepackage{epsfig} 
\usepackage{amsmath} 
\usepackage{amssymb}  
\usepackage{xcolor}
\usepackage{algorithmicx}
\usepackage[ruled, lined, linesnumbered, commentsnumbered, longend]{algorithm2e}
\usepackage{cite}
\usepackage{subcaption}
\usepackage{array}
\usepackage{hyperref}
\usepackage{subcaption}
\usepackage{caption}
\usepackage{rotating}
\DeclareCaptionLabelSeparator{periodspace}{.\quad}
\usepackage{accents}  

\def \Graph{\mathcal{G}}
\def \Factors{\mathcal{F}}

\def \Variables{\mathcal{V}}
\def \Edges{\mathcal{E}}
\def \PoseMeas{z}
\def \OdomMeas{u}
\def \PoseGround{\bar{z}}
\def \Ljoint{\mathcal{L}_{\text{joint}}}

\def \Data{\mathcal{D}}
\def \Image{\mathcal{I}}
\def \Label{y}

\DeclareMathOperator{\SE}{SE}

\DeclareMathOperator*{\argmax}{argmax}
\DeclareMathOperator*{\argmin}{argmin}

\title{\LARGE \bf
SLAM-Supported Self-Training for 6D Object Pose Estimation
}

\author{Ziqi Lu$^{1}$, Yihao Zhang$^{1}$, Kevin Doherty$^{1}$, Odin Severinsen$^{1}$, Ethan Yang$^{1}$, John Leonard$^{1}$
\thanks{*This work was supported by ONR MURI grant N00014-19-1-2571 and ONR grant N00014-18-1-2832.}
\thanks{$^{1}$ Computer Science and Artificial Intelligence Laboratory (CSAIL), Massachusetts Institute of Technology (MIT), Cambridge, MA 02139. \texttt{\{ziqilu, yihaozh, kdoherty, odinase, ethany, jleonard\}@mit.edu}.}%
}

\begin{document}

\maketitle
\thispagestyle{empty}
\pagestyle{empty}

\begin{abstract}
Recent progress in object pose prediction provides a promising path for robots to build object-level scene representations during navigation.
However, as we deploy a robot in novel environments, 
the out-of-distribution data can degrade the prediction performance.
To mitigate the domain gap, we can potentially perform self-training in the target domain, 
using predictions on robot-captured images as pseudo labels to fine-tune the object pose estimator.
Unfortunately, the pose predictions are typically outlier-corrupted, and it is hard to quantify their uncertainties, 
which can result in low-quality pseudo-labeled data.
To address the problem, we propose a SLAM-supported self-training method, 
leveraging robot understanding of the 3D scene geometry to enhance the object pose inference performance.
Combining the pose predictions with robot odometry, 
we formulate and solve pose graph optimization to refine the object pose estimates and make pseudo labels more consistent across frames.
We incorporate the pose prediction covariances as variables into the optimization to automatically model their uncertainties.
This automatic covariance tuning (ACT) process can fit 6D pose prediction noise at the component level, 
leading to higher-quality pseudo training data.
We test our method with the deep object pose estimator (DOPE) on the YCB video dataset and in real robot experiments.
It achieves respectively $34.3\%$ and $17.8\%$ accuracy enhancements in pose prediction on the two tests.
Our code is available at \url{https://github.com/520xyxyzq/slam-super-6d}.
\end{abstract}

\section{INTRODUCTION}

State-of-the-art object 6D pose estimators can capture object-level geometric and semantic information in challenging scenes \cite{sahin2020review}.
During robot navigation, an object-based simultaneous localization and mapping (object SLAM) system can use object pose predictions, with robot odometry, 
to build a consistent object-level map (\textit{e.g.} \cite{salas2013slam++}).
However, as we deploy the robot in novel environments, the pose estimators often show degraded performance on out-of-distribution data (caused by \textit{e.g.} illumination differences).
Annotating target-domain real images for training is tedious, and limits the potential for autonomous operations.
We propose to collect images during robot navigation, 
exploit the object SLAM estimates to pseudo-label the data, and fine-tune the pose estimator.

\begin{figure*}[tbh!]
    \centering
    \includegraphics[width=0.9\textwidth]{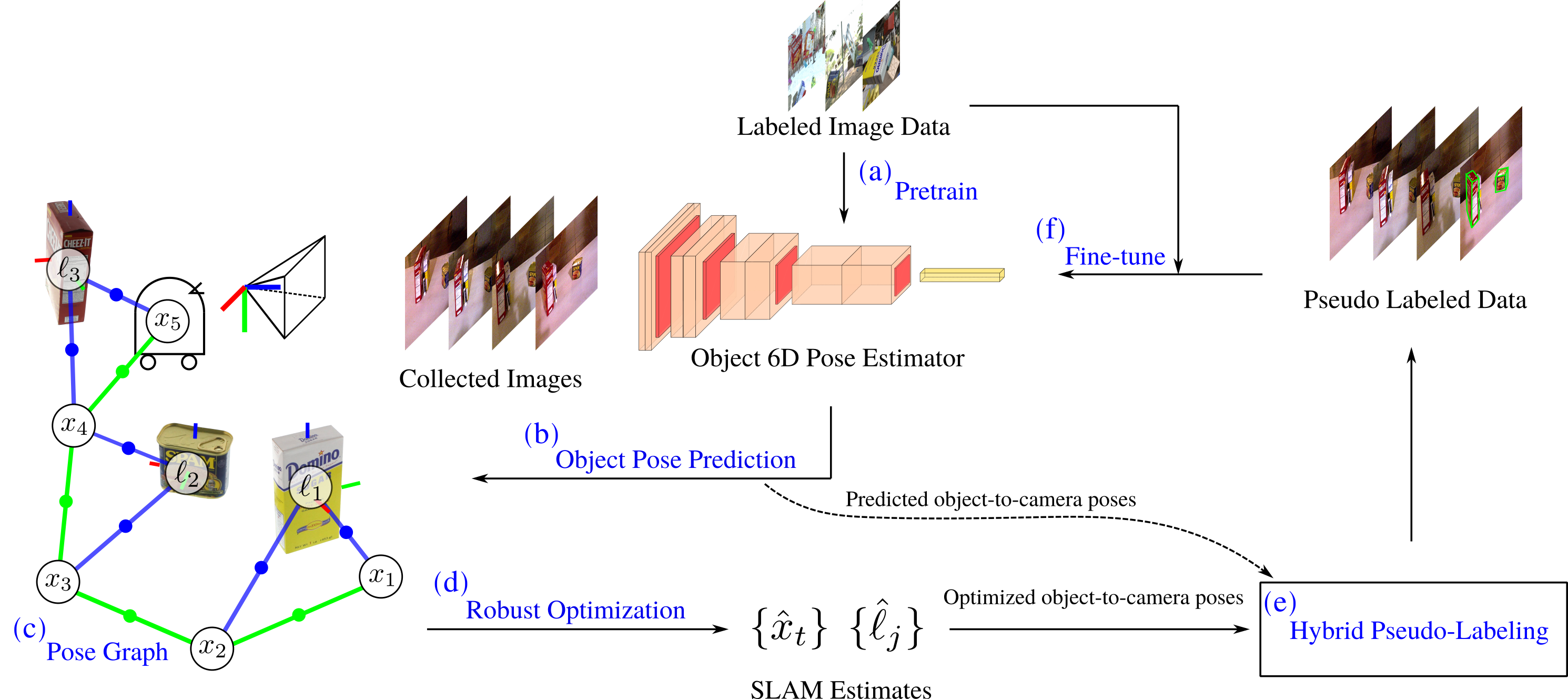}
    \caption{
    SLAM-supported self-training for object 6D pose estimators: 
    (a) Train an initial pose estimator on labeled image data; 
    (b) Infer object poses on the robot-collected RGB images; 
    (c) Establish a pose graph with object pose predictions and odometric measurements (Sec. III.A); 
    (d) Robust pose graph optimization to obtain optimized object and camera poses (Sec. III.C); 
    (e) From the optimized and predicted object-to-camera poses, 
    generate pseudo ground truth labels for the collected images via the proposed hybrid pseudo-labeling method (Sec. III.B); 
    (f) Fine-tune the initial estimator model with the pseudo-labeled data and the initial labeled data.
    }
    \label{fig1}
\end{figure*}

As depicted in Fig.~\ref{fig1}, we develop a SLAM-supported self-training procedure for RGB-image-based object pose estimators.
During navigation, the robot collects images and deploys a pre-trained model to infer object poses in the scene.
Combining the pose estimates with noisy robot state measurements from on-board odometric sensors (camera, IMU, lidar, etc.), 
a pose graph optimization (PGO) problem is formulated to optimize the camera (i.e. robot) and object poses.
We leverage the globally consistent state estimates to pseudo-label the images, 
generating new training data to fine-tune the initial model,
and mitigate the domain gap in object pose prediction without human supervision.

A major challenge in this procedure is the difficulty of modeling the uncertainty of the learning-based pose measurements \cite{abdar2021review}.
In particular, it is difficult to specify \textit{a priori} an appropriate (Gaussian) noise model for them as typically required in PGO.
For this reason, rather than fixing a potentially poor choice of covariance model, 
we allow the uncertainty model to change as part of the optimization process.
Similar approaches have been explored previously in the context of robust SLAM (e.g. \cite{agarwal2013robust, pfeifer2017dynamic}).
However, our joint optimization formulation permits a straightforward alternating minimization procedure,
where the optimal covariances are determined \textit{analytically} and \textit{component-wise}, 
allowing us to fit a richer class of noise models.

We tested our method with the deep object pose estimator (DOPE) \cite{tremblay2018deep} on the YCB video (YCB-v) dataset \cite{xiang2017posecnn} and with real robot experiments.
It can generate high-quality pseudo labels with average pixel error less than $3\%$ of the image width. 
The DOPE estimators, after self-training, make respectively $34.3\%$ and $17.8\%$ more accurate predictions, and $25.2\%$ and $25.7\%$ fewer outlier predictions in the two experiments.


In summary, our work has the following contributions:
\begin{itemize}
    \item A SLAM-aided object pose annotation method that allows robots to generate \textit{multi-view consistent} pseudo training data for object pose estimators.
    \item A robust PGO method with automatic \textit{component-wise} covariance fitting that is competitive with existing robust M-estimators, which could be of independent interest.
    \item We demonstrate that our method can effectively mitigate the domain gap for 6D object pose estimators under outlier-corrupted predictions and real sensory noise in robot navigation.
\end{itemize}

\section{Related Work}

\subsection{Semi- and self-supervised 6D object pose estimation}

Training object pose estimators, which are typically data-hungry, requires tremendous annotation efforts \cite{sahin2020review}.
While manual pose-labeling is labor-intensive, 
image synthesis tools (e.g. \cite{Morrical20nvisii}) can efficiently generate large-scale training datasets and achieve extreme realism.
However, the image rendering process still faces challenges such as realistic sensor noise simulation \cite{nikolenko2021synthetic}.

Semi- and self-supervised methods exploit unlabeled real images to mitigate the lack of data.
They typically train the pose prediction model on synthetic data in a supervised manner and improve its performance on real images by semi- or self-supervised learning \cite{zeng2017multi, li2020robust,  wang2020self6d, manhardt2020cps++, sock2020introducing, yang2021dsc, yang2021self, zhou2021semi}.
Many recent methods leverage differentiable rendering to develop end-to-end self-supervised pose estimators,
by encouraging similarity between real images and images rendered with estimated poses \cite{wang2020self6d, manhardt2020cps++, sock2020introducing, yang2021dsc}.
However, most of the methods are not leveraging
the spatial-temporal continuity of the robot-collected image streams.
Deng \emph{et al.} \cite{deng2020self}, instead, collect RGB-D image streams with a robot arm during object manipulation for self-training.
They use a pose initialization module to obtain accurate pose estimates in the first frame, 
and use motion priors from forward kinematics and a particle filter \cite{deng2019poserbpf} to propagate the pose estimates forward in time.

We propose to collect and pseudo-label RGB images during robot navigation.
Instead of frame-to-frame pose tracking, 
we directly estimate the 3D scene geometry which requires no accurate first-frame object poses or high-precision motion priors.
We use only RGB cameras and optionally other odometric sensors, which are mostly available on a robot platform, for data collection and self-training.

\subsection{Multi-view object-based perception}
Integrating information from multiple viewpoints can help correct the inaccurate and missed predictions by single-frame object perception models.
For example, Pillai \emph{et al.} \cite{pillai2015monocular} developed a SLAM-aware object localization system.
They leverage visual SLAM estimates, consisting of camera poses and feature locations, to support the object detection task, leading to strong improvements over single-frame methods.

Beyond that, multi-view object perception models can provide noisy supervision to train single-frame methods.
Mitash \emph{et al.} \cite{mitash2017self} conduct multi-view pose estimation to pseudo-label real images for self-training of an object detector.
Shugurov \textit{et al.} \cite{shugurov2021multi} leverage relative camera poses from 2 or 4 views to refine object pose estimates and fine-tune the estimator networks.
Nava \emph{et al.} \cite{nava2021uncertainty} exploit noisy state estimates to assist self-supervised learning of spatial perception models.

Inspired by these works, we propose to use SLAM-aided object pose estimation to generate training data for semi-supervised learning of object pose estimators.

\subsection{Automatic covariance tuning for robust SLAM}
Obtaining robust SLAM estimates is critical for the success of our self-training method. 
In PGO, the selection of measurement noise covariances relies mostly on empirical estimates of noise statistics.
In the presence of outlier-prone measurements (\textit{e.g.} learning-based) whose uncertainty is hard to quantify, 
it is infeasible to fix a (Gaussian) noise model for PGO.
Adaptive covariance tuning methods (\textit{e.g.} \cite{rao2011slam, vega2013cello}) have been developed to concurrently estimate the SLAM variables and noise statistics.
The adaptive noise modeling technique in general brings about more accurate SLAM estimates.

The robust M-estimators are widely applied to deal with outlier-corrupted measurements (\textit{e.g.} \cite{agarwal2013robust, pfeifer2017dynamic, yang2020graduated}).
Minimizing their robust cost functions with the iteratively re-weighted least squares (IRLS) method,
the measurement contributions are re-weighted at each step based on their Mahalanobis distances.
This down-weights the influence of outliers and is de facto re-scaling the covariance \emph{uniformly} \cite{agarwal2013robust}.
Pfeifer \emph{et al.} \cite{pfeifer2017dynamic} propose to jointly optimize state variables and covariances, with the covariance components log-regularized.
This method leads to a robust M-estimator, closed-form dynamic covariance estimation (cDCE), which can be solved with IRLS.
It is more flexible than existing M-estimators since it allows the noise components to be re-weighted \emph{differently}.

Our joint optimization formulation is similar, with covariances L1 regularized, 
and also permits closed-form component-wise covariance update.
Instead of IRLS, we follow an alternating minimization procedure,
and eliminate the influence of outliers based on the $\chi^2$ test.

\section{Methodology}

\subsection{Object SLAM via pose graph optimization}
In our method, the object pose estimator is initially trained on (typically synthetic) labeled image data 
$\Data^{l} = \{\Image^l, \PoseGround^l, \Label^l\}$,
where $\Image$ and $\PoseGround$ denote RGB images and the ground truth 6DoF object poses (w.r.t camera),
and $\Label$ represents the ground truth labels. 
We can map $\PoseGround$ to $\Label$ via an estimator-dependent function $\Label = \pi(\PoseGround)$.
In this paper, we work with the DOPE \cite{tremblay2018deep} estimator\footnote{Our method is not restricted to using a particular pose estimator.}, 
for which $\Label$ is the pixel locations for the projected object 3D bounding box, 
and $\pi(\cdot)$ is the perspective projection function.
We apply the initial estimator to make object pose predictions as the robot explores its environment.

During navigation, the robot collects a number of unlabeled RGB images $\Data^{u} = \{\Image^{u}\}$.
The estimator makes noisy object pose measurements (i.e. network predictions) $Z = \{\PoseMeas_k\in\SE(3)\}_{k=1}^{K}$ on $\Image^{u}$.
Combining $Z$ and camera odometric measurements $U = \{\OdomMeas_t\in\SE(3)\}_{t=1}^{T}$ obtained with on-board sensors,
a pose graph can be established as $\Graph \triangleq \{\Variables, \Factors, \Edges\}$ (see Fig.~\ref{fig1}(c)).
$\Variables$ denotes the latent variables incorporating camera poses $X = \{x_t\in\SE(3)\}_{t=0}^{T}$ 
and object landmark poses (w.r.t world) $L = \{\ell_j\in\SE(3)\}_{j=1}^{N}$.
Assuming all the measurements are independent and the measurement-object correspondences are known,
we can decompose the joint likelihood $p(Z, U \mid X, L)$ into factors $\Factors=\{\Factors_{U}, \Factors_{Z}\}$,
consisting of odometry factors $f_t \in \Factors_{U}$ and object pose measurement factors $f_k\in\Factors_{Z}$.
Each factor describes the likelihood of a measurement given certain variable assignments.
Edges $\Edges$ represent the factorization structure of the likelihood distribution:
\begin{equation}
\begin{aligned}
    p(Z, U \mid X, L) =& \prod_{t}f_t \prod_{k} f_k \\
    =& \prod_t p(\OdomMeas_t \mid x_{t-1}, x_{t}) \prod_k p\left(\PoseMeas_k \mid x_t, \ell_j \right)
\end{aligned}
\end{equation}
We assume the measurement noises are zero-mean, normally-distributed $\mathcal{N}(0, \Sigma_k)$, 
where $\Sigma_k$ is the noise covariance matrix for measurement $z_k$.
The optimal assignments of the variables can be obtained by solving the maximum likelihood estimation problem:
\begin{equation}
\begin{alignedat}{2}\label{mle}
    \hat{X}, \hat{L} &= \argmax_{X, L}&& \ p(Z, U \mid X, L)\\
    &= \argmin_{X, L} &&-\log p(Z, U \mid X, L)\\
    &= \argmin_{X, L} && \sum_{t}\| \text{Log}(\OdomMeas_t^{-1}x_{t-1}^{-1}x_{t}) \|^2_{\Sigma_t} +\\ 
    &&& \sum_{k}\| \text{Log}(\PoseMeas_k^{-1}x_{t}^{-1}\ell_{j}) \|^2_{\Sigma_k}
\end{alignedat}
\end{equation}
where Log$: \SE(3)\rightarrow \mathrm{R}^6$ is the logarithmic map on the SE(3) group, 
and $\|\cdot\|_{\Sigma}$ is the Mahalanobis distance.
We solve the least-squares optimization \eqref{mle} with self-tuned covariances $\Sigma_k$ (see Sec. III.C) to obtain robust SLAM estimates $\{\hat{x}_t\}, \{\hat{\ell}_j\}$.

\subsection{Hybrid pseudo-labeling}
Based on the optimal states, we can: 
(1) identify inliers $Z^{\text{in}}$ from object pose measurements $Z$;
(2) compute optimized object-to-camera poses $\hat{\PoseMeas}_{tj} = \hat{x}_t^{-1}\hat{\ell}_j$.
We compare and combine $Z^{\text{in}}$ and $\{\hat{\PoseMeas}_{tj}\}$ to obtain the pseudo ground truth poses $\bar{z}^{ps}$ for images in $\Data^u$ (see Fig.~\ref{fig1}(e)).
We refer to the pseudo-labeling method as \textit{Hybrid} labeling, because the pseudo labels are derived from two sources.

First , we identify inlier pose measurements $\PoseMeas_k^{\text{in}}\in Z^{\text{in}}\subset Z$ 
based on the $\chi^2$ test (Alg.~\ref{altmin} line \ref{chi}):
\begin{equation}\label{chi2test}
    \| \text{Log}(\PoseMeas_k^{-1}\hat{x}_{t}^{-1}\hat{\ell}_{j}) \|^2_{\Sigma_k} < \chi^2_{6, 0.95}
\end{equation}
where $\chi^2_{6, 0.95}$ is the critical $\chi^2$ value for 6 DoF at confidence level 0.95.

Second, we directly use optimized object-to-camera poses as pseudo ground truth poses.
For the image collected at time $t$, 
the optimized object-to-camera pose for object number $j$ is computed as $\hat{\PoseMeas}_{tj} = \hat{x}_t^{-1}\hat{\ell}_j$.
However, the inlier predictions or optimized poses can be noisy and may deviate visually from the target objects on the images and hurt training.

Therefore, we employ a pose evaluation module, as proposed in \cite{deng2020self}, 
to quantify the visual coherence between pseudo labels and images.
Each RGB image is compared with its corresponding rendered image based on the pseudo ground truth pose $\PoseGround^{ps}$.
The regions of interest (RoIs) on the two images are passed into a pre-trained auto-encoder from PoseRBPF \cite{deng2019poserbpf},
and the cosine similarity score of their feature embeddings is computed.
We abstract the score calculation process as a function $\phi: \SE(3)\rightarrow\left[0, 1\right]$, 
where the dependence on object, camera and image information is made implicit.

We compute the PGO-computed pose's score $\phi(\hat{\PoseMeas})$ and the inlier score $\phi(z^{\text{in}})$ for every target object on unlabeled images,
and pick the pose with a higher score as the pseudo ground truth pose $\bar{z}^{ps}$ (see Sec. IV.A for details).
We assemble the pseudo-labeled data as $\Data^{ps} = \{\Image^{ps}, \PoseGround^{ps}, \Label^{ps}\}$,
where again $\Label^{ps} = \pi(\PoseGround^{ps})$.
And we fine-tune the object pose estimator with $\Data^{ps}\cup\Data^{l}$ (see Fig.\ref{fig1}(f)).

Our method naturally generates training samples on which the initial estimator fails to make reasonable predictions, i.e. \textit{hard examples}.
In particular, when the initial estimator makes an outlier prediction or misses a prediction,
PGO can potentially recover an object pose that is visually consistent with the RGB image.
The challenging examples, as we demonstrate in Sec. IV, are the key to achieving significant performance gain during self-training.

\subsection{Automatic covariance tuning by alternating minimization}
Since the object pose measurements $Z$ are derived from learning-based models,
it is difficult to \textit{a priori} fix a (Gaussian) uncertainty model for them as typically required in PGO.
We propose to automate the covariance tuning process by incorporating the noise covariances as variables into a joint optimization procedure.
Our formulation, as we show, leads to a simple and robust PGO method,
that allows us to alternate between PGO and determining \textit{in closed form} the optimal noise models.

For simplicity, we rewrite the PGO cost function in \eqref{mle} as:
\begin{equation}
    \mathcal{L} = \sum_{k} \|e_k\|_{\Sigma_k}^{2} + \sum_{t} \|e_{t}\|_{\Sigma_{t}}^{2}
    \label{pgoloss}
\end{equation}
where $e_k$ and $e_t$ are the residual error vectors.

Given that the odometry measurements $U$ are typically more reliable,  
we only update the object pose noise covariances $\Sigma_k$ at the covariance optimization step.
Each time after optimizing SLAM variables, 
we solve for the per-measurement noise models $\Sigma_k$ that can further minimize the optimization loss, and re-solve the SLAM variables with them.
To avoid the trivial solution of $\Sigma_k \to \infty$, we apply L1 regularization on the covariance matrices. 
The loss function for the joint optimization is:
\begin{equation}
    \Ljoint = \sum_{k} \|e_k\|_{\Sigma_k}^{2} + \lambda\sum_{k} \|\Sigma_k\|_1 + \sum_{t} \|e_{t}\|_{\Sigma_{t}}^{2}
    \label{joint}
\end{equation}

For simplicity, we assume the covariance matrices are diagonal, i.e. $\Sigma_k = \text{diag}(\sigma^2_{k1}, \cdots, \sigma^2_{k6})$, and the joint loss reduces to:
\begin{equation}
    \Ljoint = \sum_{k} \sum_{j=1}^{6} \frac{e_{kj}^2}{\sigma_{kj}^2} + \lambda\sum_{k}\sum_{j=1}^{6} \sigma_{kj}^2 + \sum_{t} \|e_{t}\|_{\Sigma_{t}}^{2}
    \label{joint_scalar}
\end{equation}
where $e_{kj}$ is the $j$th entry of the residual error $e_k$. 

For fixed $e_t$ and $e_{kj}$, optimizing \eqref{joint_scalar} w.r.t $\sigma^2_{kj}$ amounts to minimizing a convex objective over the non-negative reals. 
Moreover, we know $\sigma^2_{kj}=0$ is never a minimizer for $\mathcal{L}_{\text{joint}}$.
Thus, a global optimal solution is obtained when:
\begin{equation}
    \frac{\partial \Ljoint}{\partial \sigma_{kj}^2} = - \frac{e_{kj}^2}{\sigma_{kj}^{4}} + \lambda = 0
\end{equation}
evaluating which we obtain the extremum on $\mathcal{L}_{\text{joint}}$:
\begin{equation}\label{extremum}
    {\sigma^2_{kj}} = \frac{|e_{kj}|}{\sqrt{\lambda}} \quad j = 1, \cdots, 6
\end{equation}
With $\partial^2  \mathcal{L}_{\text{joint}}/ \partial \sigma_{kj}^2  \geq 0$ being valid for all $\sigma^2_{kj}>0$, 
we can confirm that this extremum is a global minimum.

Therefore we can express the covariance update rule at iteration $i$ as:
\begin{equation}\label{update_rule}
    \Sigma_k^{(i)} = \text{diag}(\lambda ' \lvert e_k^{(i)}\rvert)
\end{equation}
where we define $\lambda' =1/\sqrt{\lambda}$ for the convenience of $\lambda$ tuning.

Since the covariance optimization step admits a closed form solution,
the joint optimization reduces to ``iteratively solving PGO with re-computed covariances".
The selection of all the noise models reduces to tuning $\lambda$.
According to our experiments, 
it's feasible to set $\lambda$ as a constant for consistent performance in different cases\footnote{$\lambda'=10$ (i.e. $\lambda=0.01$) for all the tests.}.

The algorithm is summarized in Alg.~\ref{altmin}.
We use the Levenberg-Marquardt (LM) algorithm to solve the PGO with Gaussian noise models (line \ref{lm}).
For better performance, we also identify outliers at each iteration using the $\chi^2$ test \eqref{chi2test} and set their noises to a very large value (line \ref{explode}) to rule out their influences.
The optimization is terminated as the relative decrease in $\mathcal{L}_{\text{joint}}$ is sufficiently small or the maximum number of iterations is reached.
In Appendix \textit{A}, we show the algorithm can monotonically improve $\mathcal{L}_{\text{joint}}$.

\begin{algorithm}[htb]
    \scriptsize
    \caption{ACT by alternating minimization}
    \label{altmin}
    \KwIn{Pose graph $\Graph \triangleq \{ \{\Factors_Z, \Factors_U\}, \Variables, \Edges\}$ with object pose
    measurement factors $f_k\in\Factors_{Z}$ and odometry factors $f_t\in\Factors_{U}$. \ Initial values: $\Variables^{(0)}$, $\{\Sigma_k^{(0)}\}$, $\{\Sigma_{t}^{(0)}\}$. \ Regularization coefficient: $\lambda '$.}
    \KwOut{Optimized state variables $\Variables^{*}$}

    \SetKwFunction{LM}{LM}
    \SetKwFunction{evaluateError}{evaluateError}
    \SetKwFunction{chiSqTest}{chiSqTest}
    \For {$i\leftarrow 1$ \KwTo $N$}{
        $\Variables^{(i)}$ = $LM(\Graph$, $\Variables^{(i-1)}$, $\{\Sigma_k^{(i-1)}\}$, $\{\Sigma_{t}^{(0)}\})$\hfill\Comment{solve PGO by LM} \label{lm}\\
        \For{$k\leftarrow 1$ \KwTo $K$}{
            $e_k^{(i)} = evaluateError(f_k(\Variables^{(i)}))$ \hfill\Comment{residual error vector}\\
            \eIf{$chi^2Test(e_k^{(i)}, \Sigma_k^{(0)})$}{\label{chi}
                $\Sigma_k^{(i)} = \text{diag}(\lambda ' \lvert e_k^{(i)}\rvert)$\hfill\Comment{rescale cov. using error comp.} \\ \label{update}
            }{
                $\Sigma_k^{(i)} = 10^{10}\mathcal{I}$ \hfill\Comment{remove outliers' influence}\\ \label{explode}
            }
        }
    }
    \Return{$\Variables^{(N)}$}
\end{algorithm}

The update rule \eqref{update_rule} reveals our implicit assumption that 
a high-residual measurement component is likely from a high-variance normal distribution.
The recomputed covariances down-weight high-residual measurements, which is in spirit similar to robust M-estimation.
We show in Appendix \textit{B} that, with isotropic noise models, our method reduces to using the L1 robust kernel.

However, the robust M-estimation is typically solved with the IRLS method, 
where the measurement losses are re-weighted based on the Mahalanobis distance $\|e_k\|_{\Sigma_k}$ (see \eqref{irls}).
In comparison, our method re-weights the measurement losses component-wise using residual error components.
This enables us to fit a richer class of noise models for 6DoF PGO, 
in that different components often follow different noise characteristics.


\section{Experiments}
Our method is tested with (1) the YCB-v dataset and (2) a real robot experiment.
On the YCB-v dataset (Sec. IV.A), 
we leverage the image streams in the training sets to solve per-video PGOs for self-training.
We also conduct ablation experiments to compare the \textit{Hybrid} labeling method against another two baseline methods.
In the real robot experiment (Sec. IV. B), 
we apply the method on longer sequences, circumnavigating selected objects, and test our method's robustness to challenges in ground robot navigation, \textit{e.g.} increased motion blur.

Our method is implemented in Python.
We use the NVISII toolkit \cite{Morrical20nvisii} to generate synthetic image data.
The training and pose inference programs are adapted from the code available in the DOPE GitHub \cite{tremblay2018deep}.
Every network is initially trained for 60 epochs and fine-tuned for 20 epochs, 
with a batchsize of 64 and learning rate of 0.0001.
We solve PGO problems based on the GTSAM library \cite{dellaert2012factor}.
We run the synthetic data generation and training programs with 2 NVIDIA Volta V100 GPUs 
and other code on an Intel Core i7-9850H@2.6GHz CPU and an NVIDIA Quadro RTX 3000 GPU.

\subsection{YCB video experiment}
\begin{figure}[tb!]
    \centering
    \begin{subfigure}{0.23\textwidth}
        \centering
        \includegraphics[width=\textwidth]{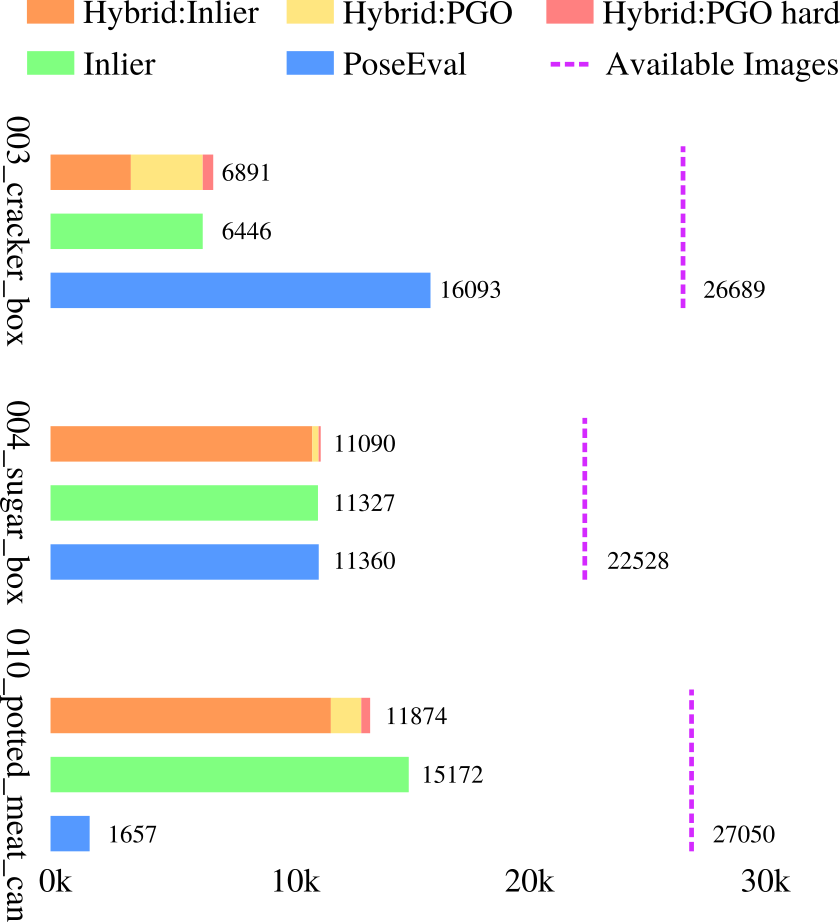}
        \caption{Data sizes}
    \end{subfigure}%
    \begin{subfigure}{0.25\textwidth}
        \centering
        \includegraphics[trim=0 0 60 0, width=\textwidth]{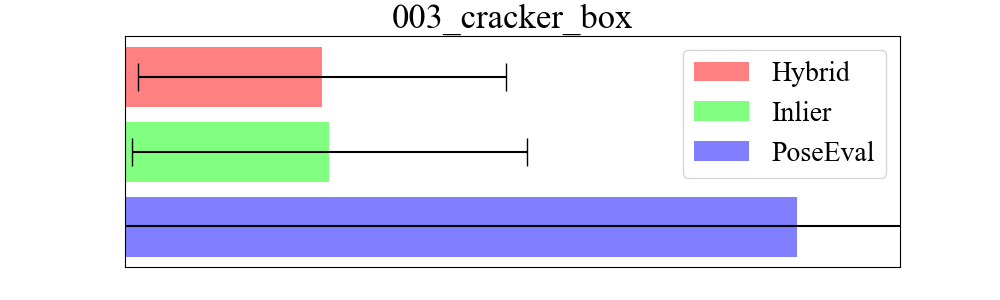}
        \includegraphics[trim=0 0 60 0, width=\textwidth]{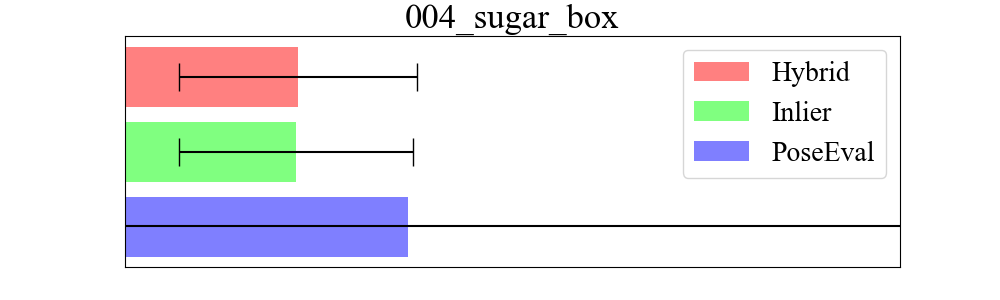}
        \includegraphics[trim=0 0 60 0, width=\textwidth]{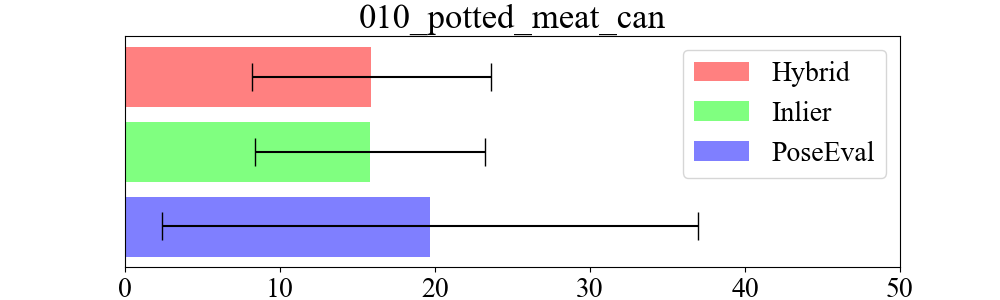}
        \caption{Label errors in pixels}
    \end{subfigure}
    \caption{\textbf{Statistics of the pseudo-labeled data} generated on the YCB-v training set.
        Our \textit{Hybrid} pseudo-labeled data are derived from 2 sources: 
        inlier pose predictions and PGO-estimated poses, and the latter consists of easy and hard examples.
        \textit{Inlier} and \textit{PoseEval} are baseline labeling methods.
        (a) Data sizes; 
        (b) Average pixel errors in the pseudo labels (Image size: $480\times640$). 
        The black bars represent $\pm$ one standard deviation.
    }
    \label{stats}
\end{figure}

We tested our method with the DOPE estimators \cite{tremblay2018deep} for three YCB objects: 
003\_cracker\_box, 004\_sugar\_box and 010\_potted\_meat\_can.
Each object appears in 20 YCB videos (training + testing).
They have respectively 26689, 22528 and 27050 training images (dashed lines in Fig.~\ref{stats}(a)) from 17, 15, and 17 training videos.

We use 60k NVISII-generated synthetic image data $\Data^l$ to train initial DOPE estimators for the 3 objects\footnote{The 010\_potted\_meat\_can data are the same as that used in \cite{Morrical20nvisii}.}.
The models are applied to infer object poses on the YCB-v images.
We employ the ORB-SLAM3 \cite{campos2021orb} RGB-D module (w/o loop closing) to obtain camera odometry on these videos.

\begin{table}[tb]
    \tiny
    \centering
    \noindent
    \caption{
        \textbf{Comparison of robust PGO methods via pseudo label accuracy} on YCB-v sequences. 
        Column 1-7 are median (pixel) errors in PGO-generated pseudo labels $\Label_{tj}$ for the first 7 (out of 20) videos. 
        Column 8 (\#best) is the number of sequences on which a method achieves the lowest error.
    }
    \begin{tabular}{c | c c c c c c c >{\color{blue}}c}
        \hline
        003\_cracker\_box & 0001 & 0004 & 0007 & 0016 & 0017 & 0019 & 0025 & \#best  \\
        \hline
        LM & 62.3 & 58.7 & 13.2 & 69.4 & 37.6 & 110.1 & 101.6 & 0 \\
        Cauchy & 12.4 & \textbf{10.8} & 10.2 & 13.8 & 29.5 & 94.4 & 171.4 & 4 \\
        Huber & 31.4 & 25.4 & 10.2 & 34.2 & 21.6 & 52.5 & 57.0 & 1 \\
        Geman-McClure & \textbf{11.5} & 168.4 & 10.2 & 115.0 & 48.4 & 94.4 & 171.4 & 2\\
        cDCE\cite{pfeifer2017dynamic} & 28.7 & 25.4 & 10.5 & 32.5 & 21.1 & \textbf{45.2} & 58.9 & 4\\
        ACT(Ours) &  15.7 & 12.0 & \textbf{9.4} & \textbf{12.6} & \textbf{20.3} & 52.0 & \textbf{15.4} & \textbf{9}\\
        \hline
        004\_sugar\_box & 0001 &  0014 &  0015 & 0020 & 0025 & 0029 & 0033  & \#best\\
        \hline
        LM & 22.9 & 27.1 & 100.9 & 21.1 & 57.3 & 78.7 & 7.1 & 0 \\
        Cauchy & 8.3 & 13.4 & 30.4 & 21.9 & 22.3 & 104.4 & 6.4  & 1 \\
        Huber & 11.7 & 12.8 & 35.5 & 15.8 & 23.4 & \textbf{71.1} & 6.6  & 3 \\
        Geman-McClure & 9.4 & \textbf{11.4} & \textbf{29.1} & \textbf{14.3} & 19.6 & 104.4 & 6.4  & 6 \\
        cDCE\cite{pfeifer2017dynamic} & 12.3 & 12.0 & 31.6 & 14.9 & 20.0 & 72.7& 6.5  & 0 \\
        ACT(Ours) &  \textbf{8.2} & 15.9 & 34.2 & 15.1 & \textbf{18.0} & 100.5 & \textbf{6.1} & \textbf{10} \\
        \hline
        010\_potted\_meat\_can & 0002 & 0005 & 0008 & 0014 & 0017 & 0023 & 0026 & \#best  \\
        \hline
        LM & 35.2 & 38.1 & 61.4 & 59.2 & 31.1 & 32.8 & 17.5 & 0 \\
        Cauchy & 10.8 & 14.9 & \textbf{10.7} & 12.2 & \textbf{14.2} & \textbf{11.7} & 11.4 & 6 \\
        Huber & 11.1 & 17.1 & 14.6 & 16.6 & 18.6 & 15.5 & 11.8 & 1\\
        Geman-McClure & \textbf{10.4} & 15.3 & 11.9 & 13.2 & 18.9 & 15.2 & \textbf{9.1}& 5\\
        cDCE\cite{pfeifer2017dynamic} & 11.5 & 16.2 & 16.1 & 20.5 & 17.8 & 15.3 & 11.2 & 1 \\
        ACT(Ours) &  13.3 & \textbf{14.5} & 12.8 & \textbf{11.3} & 19.1 & 14.0 & 10.4 & \textbf{7}\\
        \hline
    \end{tabular}
    \label{keypt_err}
\end{table}

Combining the measurements, we solve per-video PGOs to refine object-to-camera poses for pseudo-labeling.
We initialize the camera poses with the odometry chain, 
object poses with average pose predictions,
and covariances with $\Sigma_k = 0.1\mathcal{I}, \Sigma_t = 0.01\mathcal{I}$.
We apply different robust optimization methods to solve all 60 PGO problems (see Tab.~\ref{keypt_err}\footnote{
    Due to space limitations, we report results for the first 7 out of 20 YCB-v sequences.
    Please check out our GitHub repo for complete statistics.
}\footnote{ We use the default parameters and the same noise covariances for the robust M-estimators. 
}\footnote{
    We implemented cDCE via replacing \eqref{extremum} with equation (16) in \cite{pfeifer2017dynamic}.
}).
For comparison purposes only, we compute pseudo labels for all the YCB-v images directly from the optimal states, 
i.e. $\Label_{tj}=\pi(\hat{x}_t^{-1}\hat{\ell}_j)$,
and compare the methods via label errors, 
i.e. how much the projected object bounding boxes deviate from the ground truth.
With the component-wise noise modelling and mitigation of outlier corruption,
our method (ACT) achieves the lowest errors in much more videos (see Tab.~\ref{keypt_err} Col. 8).
It performs stably across sequences based on a fixed initial guess and a constant regularization coefficient $\lambda$.
Therefore, we elect to pseudo-label the YCB-v training images based on the results by ACT.

To evaluate different modules, 
we compare our \emph{Hybrid} method with another two baseline labeling methods: \textit{Inlier} and \textit{PoseEval}.
\textit{Inlier} uses the PGO results only as an inlier measurement filter.
The raw pose predictions that agree geometrically with other measurements are selected for labeling,
i.e. $\PoseGround^{ps} = Z^{\text{in}}$.
\textit{PoseEval} extracts visually coherent pose predictions by thresholding the similarity scores from the pose evaluation module,
i.e. $\PoseGround^{ps} = \{\PoseMeas_k \mid \phi(\PoseMeas_k) > s^{\text{pe}}_{*}\}$\footnote{$s^{\text{pe}}_{*}=0.5$ as in \cite{deng2020self}.}.

The \textit{Hybrid} method ensures spatial \textit{and} visual consistency of the pseudo-labeled data. 
For a certain target object in an image, 
we compare the pose evaluation score for the inlier prediction $\phi(\PoseMeas_k^{\text{in}})$ (if available) 
with that of the optimized object pose $\phi(\hat{\PoseMeas})$.
The higher scorer, if beyond a threshold, is picked for labeling. 
Thus, the \textit{Hybrid} pseudo ground truth poses can be expressed as: 
$\PoseGround^{ps} = \{\hat{\PoseMeas}\mid\phi(\hat{\PoseMeas}) > \phi(\PoseMeas_k^{\text{in}}) \land \phi(\hat{\PoseMeas}) > s^{\text{pgo}}_{*}\} \cup \{\PoseMeas_k^{\text{in}}\mid\phi(\PoseMeas_k^{\text{in}}) > \phi(\hat{\PoseMeas}) \land \phi(\PoseMeas_k^{\text{in}}) > s^{\text{in}}_{*}\}$,
where the thresholds satisfy $s^{\text{pgo}}_{*} > s^{\text{in}}_{*}$ 
because the PGO-generated labels, not directly derived from RGB images, are prone to misalignment\footnote{
$s^{\text{pgo}}_{*} = 0.9, 0.8, 0.5$ and $s^{\text{in}}_{*} = 0.3, 0.3, 0.2$ for the 3 YCB objects.
}.
Thus, the \textit{Hybrid} pseudo-labeled data $\Data^{ps}$ consists of high-score inliers, PGO-generated easy examples,
and hard examples (on which the initial estimator fails).
The 3 components are colored differently on \textit{Hybrid} bars in Fig.~\ref{stats}(a).
For the \textit{Hybrid} and \textit{Inlier} modes, 
we also exclude YCB-v sequences with measurement outlier rates higher than 20\% from pseudo-labeling.

\begin{figure*}[htb!]
    \centering
    \begin{sideways}
        \begin{minipage}[c]{0.05\textwidth}
            \caption*{Before}
        \end{minipage}
    \end{sideways}
    \vspace{1mm}
    \begin{subfigure}{0.18\linewidth}
        \centering
        \includegraphics[trim=0 60 0 45, clip, width=\linewidth]{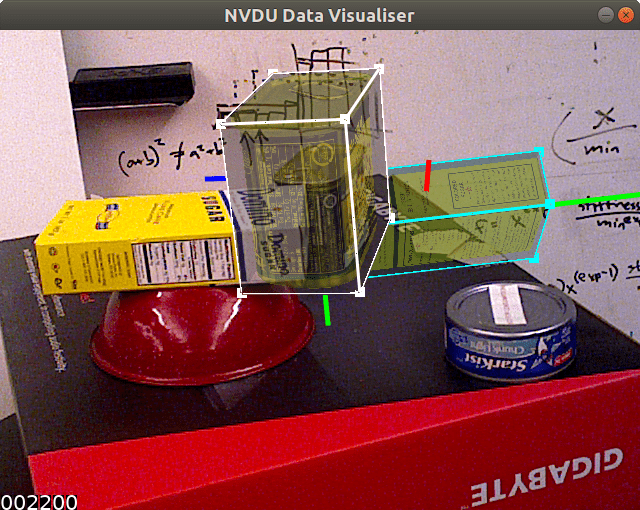}
    \end{subfigure}
    \begin{subfigure}{0.18\linewidth}
        \centering
        \includegraphics[trim=0 60 0 45, clip, width=\linewidth]{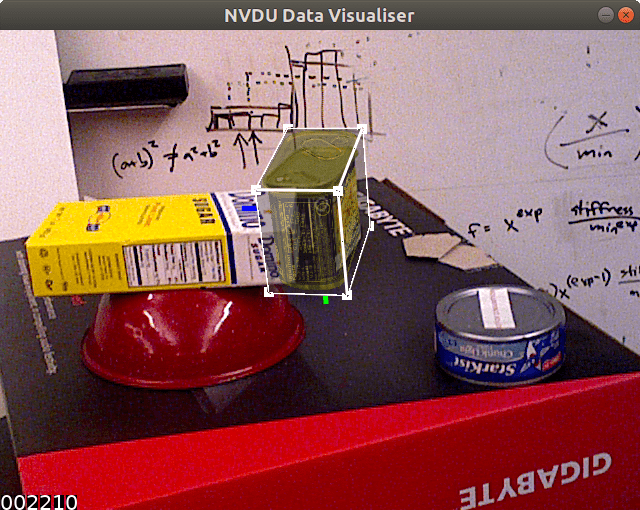} 
    \end{subfigure}
    \begin{subfigure}{0.18\linewidth}
        \centering
        \includegraphics[trim=0 60 0 45, clip, width=\linewidth]{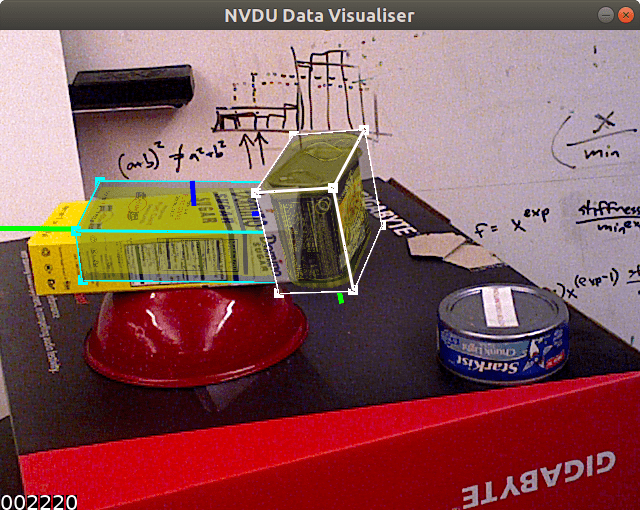}  
    \end{subfigure}
    \begin{subfigure}{0.18\linewidth}
        \centering
        \includegraphics[trim=0 60 0 45, clip, width=\linewidth]{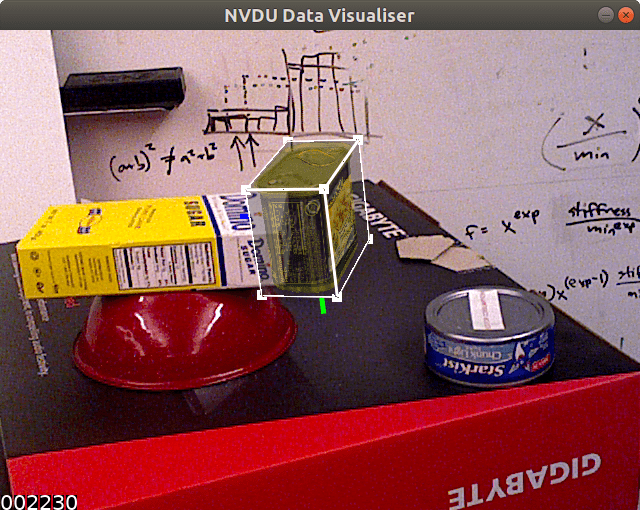}
    \end{subfigure}
    \begin{subfigure}{0.18\linewidth}
        \centering
        \includegraphics[trim=0 60 0 45, clip, width=\linewidth]{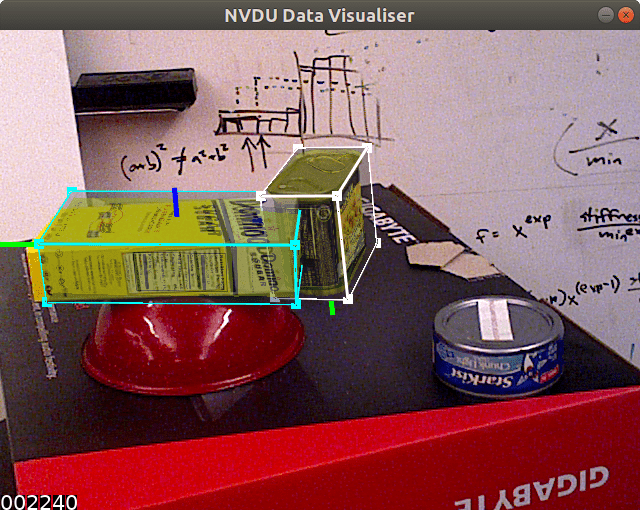}
    \end{subfigure}
    \\
    \begin{sideways}
        \begin{minipage}[c]{0.05\textwidth}
            \caption*{After}
        \end{minipage}
    \end{sideways}
    \begin{subfigure}{0.18\linewidth}
        \centering
        \includegraphics[trim=0 60 0 45, clip, width=\linewidth]{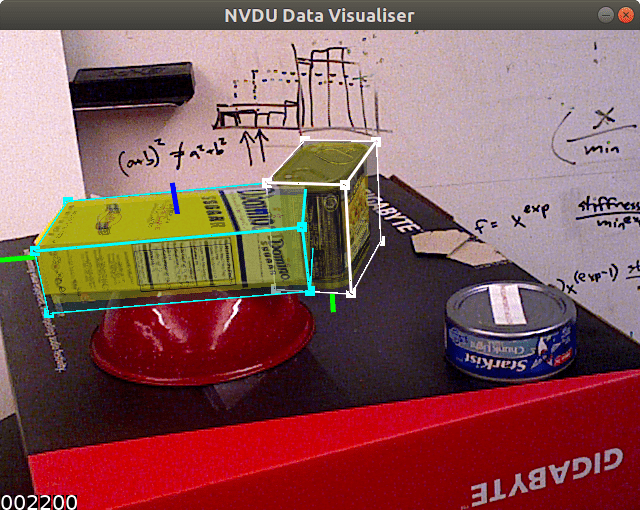}
        \caption*{$t=1$}          
    \end{subfigure}
    \begin{subfigure}{0.18\linewidth}
        \centering
        \includegraphics[trim=0 60 0 45, clip, width=\linewidth]{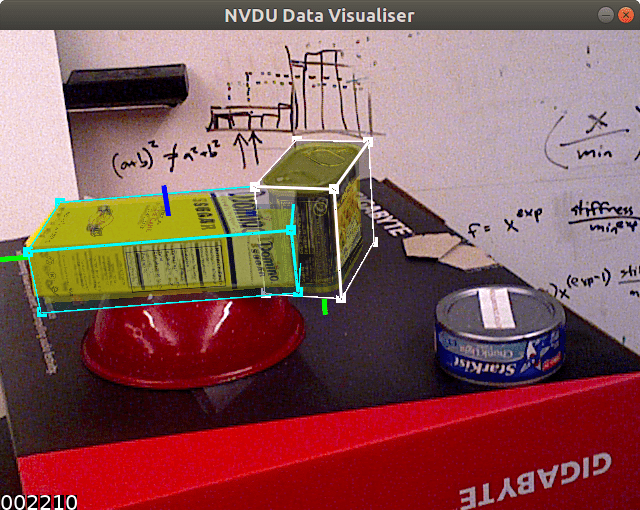}
        \caption*{$t=2$}        
    \end{subfigure}
    \begin{subfigure}{0.18\linewidth}
        \centering
        \includegraphics[trim=0 60 0 45, clip, width=\linewidth]{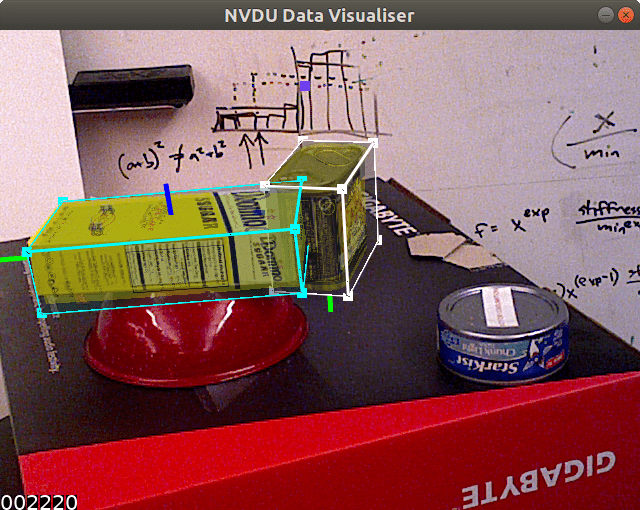}
        \caption*{$t=3$}
    \end{subfigure}
    \begin{subfigure}{0.18\linewidth}
        \centering
        \includegraphics[trim=0 60 0 45, clip, width=\linewidth]{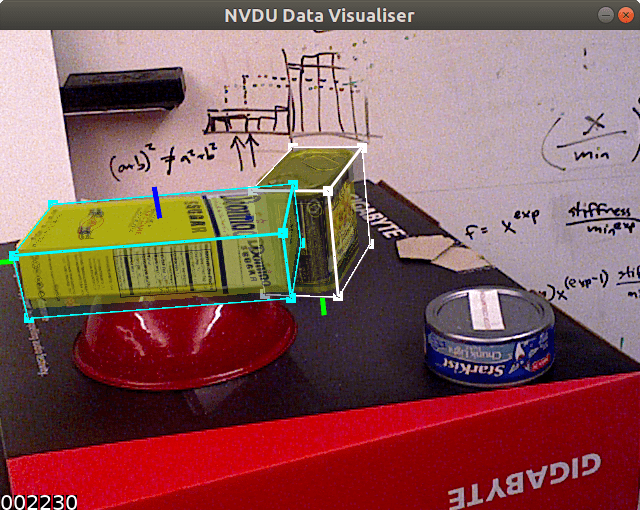}
        \caption*{$t=4$}
    \end{subfigure}
    \begin{subfigure}{0.18\linewidth}
        \centering
        \includegraphics[trim=0 60 0 45, clip, width=\linewidth]{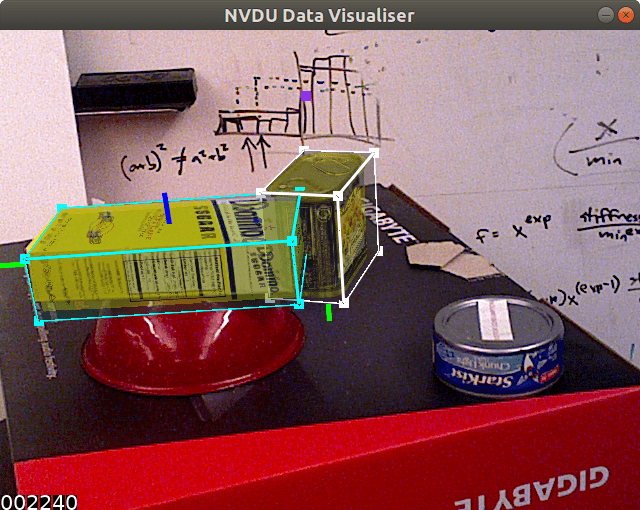}
        \caption*{$t=5$}
    \end{subfigure}
    \caption{
        \textbf{Qualitative results} on the YCB-v test set 0049: 
        per-frame object pose predictions by the initial (top row) and self-trained (bottom row) estimators. 
        The predictions by the self-trained estimator are more consistent across frames and have fewer outliers.
        (Data visualization by NVDU \cite{nvdu2018nvdu}.)
    }
    \label{qual}
\end{figure*}

The statistics of the pseudo-labeled data are reported in Fig.~\ref{stats}.
The \textit{Hybrid} and \textit{Inlier} data are in general very accurate (average label errors $<3\%$ of image width), 
with the $\chi^2$ test being an effective inlier filter.
But the \textit{PoseEval} data, whatever their sizes, are more noisy and outlier-corrupted,
so we cannot rely only on the pose evaluation test to generate outlier-free labels.

\begin{figure}[htb!]
    \centering
    \begin{subfigure}{0.333\linewidth}
        \centering
        \includegraphics[width=\linewidth]{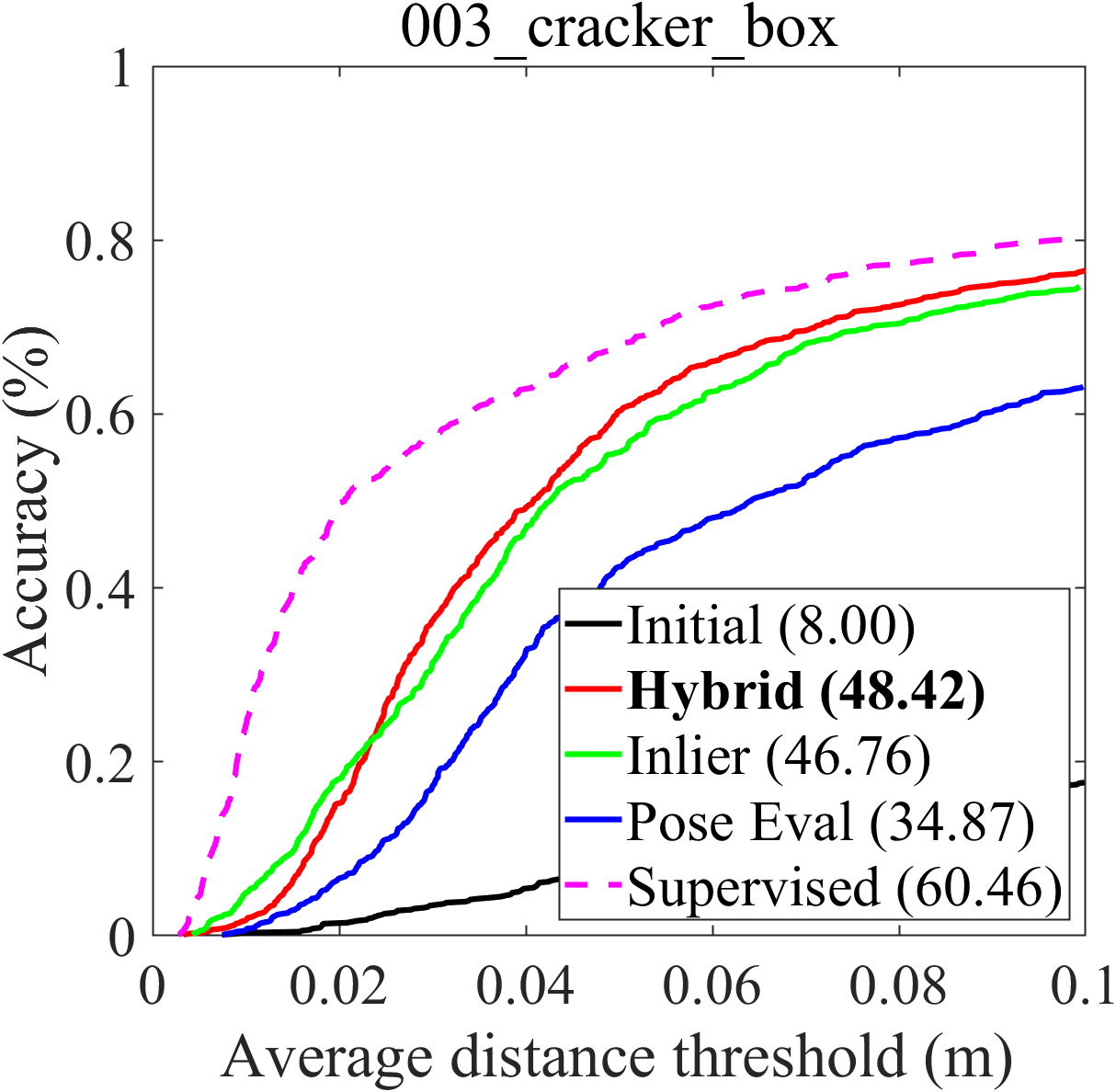}
    \end{subfigure}%
    \begin{subfigure}{0.333\linewidth}
        \centering
        \includegraphics[width=\linewidth]{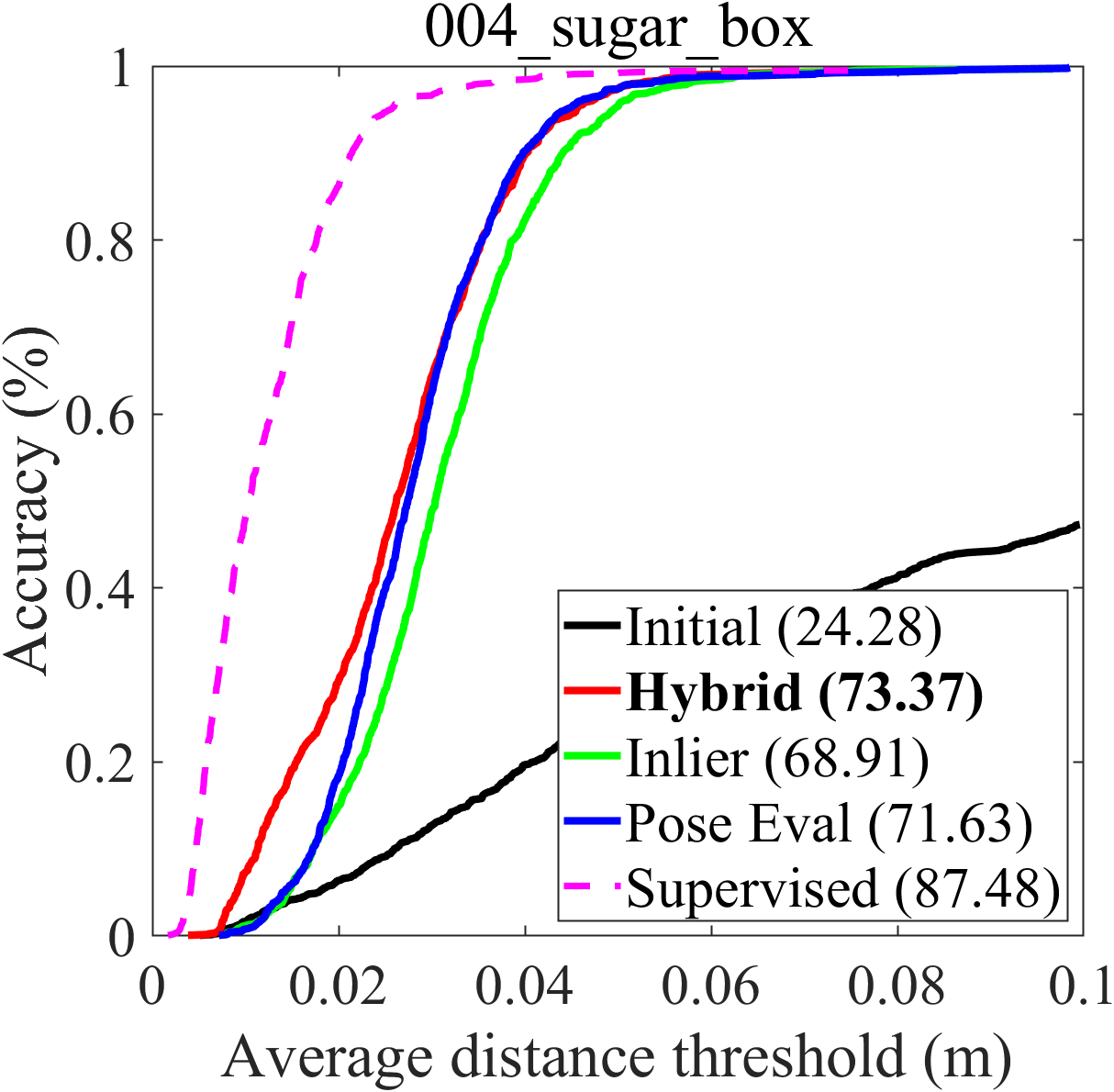}
    \end{subfigure}%
    \begin{subfigure}{0.333\linewidth}
        \centering
        \includegraphics[width=\linewidth]{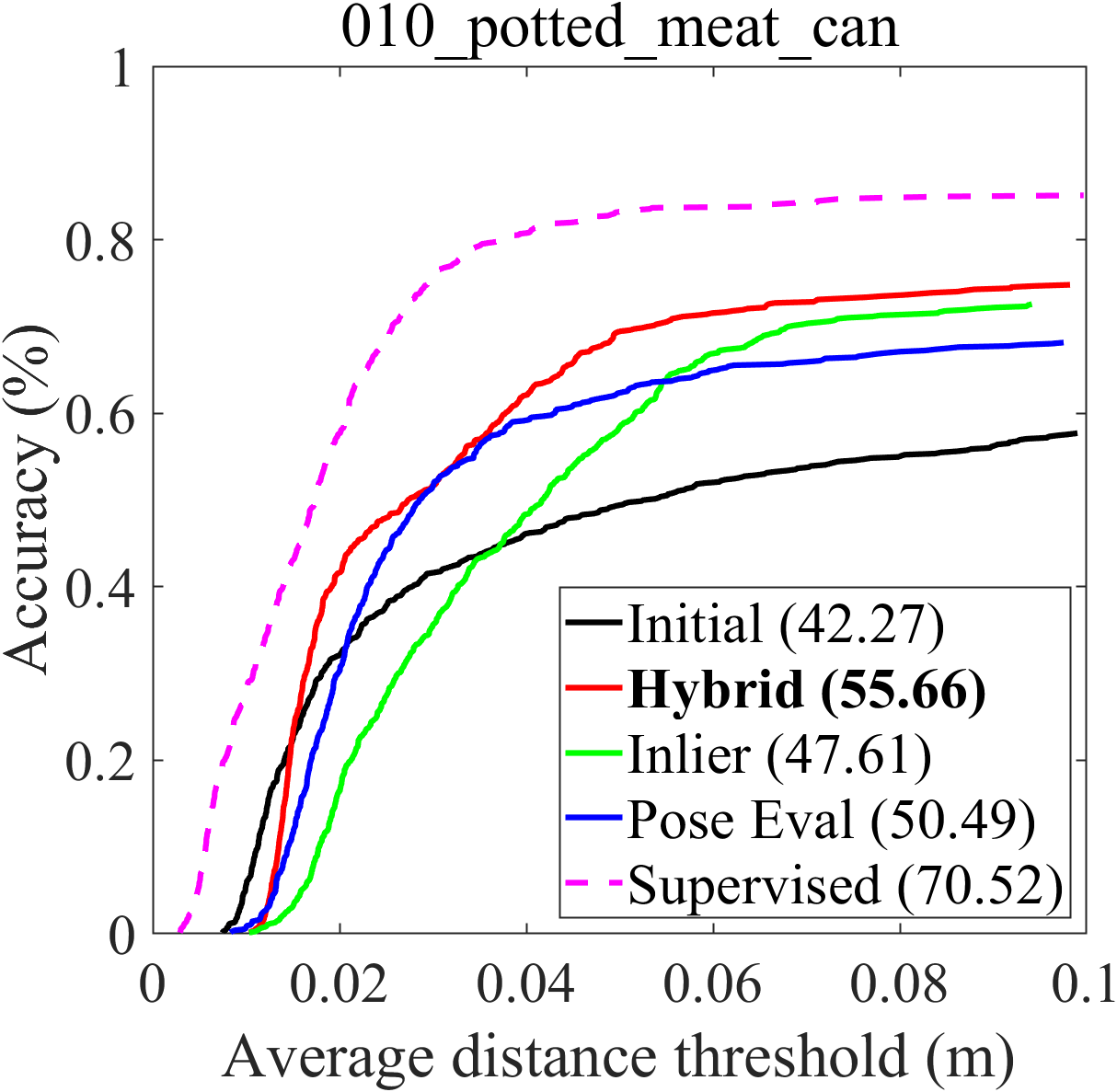}
    \end{subfigure}%
    \caption{
    \textbf{Quantitative evaluation} of self-training methods on the YCB-v test sets: ADD accuracy-threshold curves before (\textit{Initial}) and after self-training. 
    The area under curve (AUC) values (\%) are labeled in the legend.
    Our self-trained estimator (\textit{Hybrid}) achieves on average \textbf{34.3\%} more accurate predictions over the initial pose estimator.
    }
    \label{ycbv_eval}
\end{figure}

We evaluate different self-training methods, using the estimator performance gain, on the YCB-v test sets.
We adopt the average distance (ADD) metric \cite{hinterstoisser2012model}, 
and present the accuracy-threshold curves in Fig.~\ref{ycbv_eval}.
All the methods achieve considerable improvements over the initial model,
indicating that the real image data, although noisy and outlier-corrupted, 
are highly effective for decreasing the domain gap, 
as also reported in \cite{wang2020self6d, sock2020introducing, manhardt2020cps++, zhou2021semi, yang2021dsc}.
But they still have large performance gaps from the models trained by means of supervised learning, 
due to noises and the limited data size.
Our \textit{Hybrid} method consistently outperforms other baselines, 
even though its data have similar statistics with the \textit{Inlier} data.
We thus infer that the performance gain mainly comes from the presence of hard examples.

We also present the performance gain qualitatively in Fig.~\ref{qual}, and the outlier prediction reduction after self-training in Tab.~\ref{outreduce}.
We observe that the per-frame pose predictions become more consistent across frames and the outlier predictions are minimized (-25.2\%) after self-training, even if the initial predictions are outlier-contaminated.

Further, we examine whether the fine-tuned estimator brings about improved object SLAM accuracy.
The YCB-v test sequences 0049, 0059 are selected for evaluation 
(since they have 2 out of the 3 selected YCB objects).
We solve the PGOs with the LM algorithm, using object pose predictions before and after the estimator is fine-tuned.
The estimation errors are reported in Tab.~\ref{slam_compare_tab}.
We observe that camera and object pose estimations are consistently more accurate after model self-training.

\begin{table}[htb!]
    \centering
    \caption{ The \textbf{outlier prediction rates} on the YCB-v test sets before and after self-training. We use the $\chi^2$ test to identify outliers from all the object pose predictions.
    }
    \begin{tabular}{c | c c}
        \hline
        \% Outliers            & Before & After\\
        \hline
        003\_cracker\_box      & 58.43   & \textbf{12.07}\\
        004\_sugar\_box        & 21.76   & \textbf{0.27}\\
        010\_potted\_meat\_can & 15.06   & \textbf{2.29}\\
        \hline
        Total                  & 29.07   & \textbf{3.88}\\
        \hline
    \end{tabular}
    \label{outreduce}
\end{table}

\begin{table}[htb!]
    \centering
    \scriptsize
    \caption{
        \textbf{Object SLAM accuracy} on the YCB-v test sequences 0049 and 0059 \textbf{before and after self-training}: 
        absolute trajectory errors (ATE) and object pose errors (w.r.t world). 
    }
    \begin{tabular}{c | c | c c  c c}
        \hline
        \multicolumn{2}{c|}{0049} & \multicolumn{2}{c}{004\_sugar\_box} & \multicolumn{2}{c}{010\_potted\_meat\_can}\\
        \hline
               & ATE (m) & Tran. (m)       & Ori (rad)      & Tran. (m)       & Ori (rad) \\
        \hline
        Before & 0.199          & 0.054          & \textbf{0.023} & 0.091          & 0.154 \\
        After  & \textbf{0.081} & \textbf{0.039} & 0.083          & \textbf{0.037} & \textbf{0.089} \\
        \hline
        \multicolumn{2}{c|}{0059} & \multicolumn{2}{c}{003\_cracker\_box} & \multicolumn{2}{c}{010\_potted\_meat\_can}\\
        \hline
               & ATE (m) & Tran. (m)       & Ori (rad)      & Tran. (m)       & Ori (rad) \\
        \hline
        Before & 0.618          & 0.234          & 0.402          & 0.518          & 0.214 \\
        After  & \textbf{0.105} & \textbf{0.057} & \textbf{0.027} & \textbf{0.032} & \textbf{0.147} \\
        \hline
    \end{tabular}
    \label{slam_compare_tab}
\end{table}

\subsection{Real robot experiment}
\begin{figure}[htb!]
    \centering
    \includegraphics[width=0.7\linewidth]{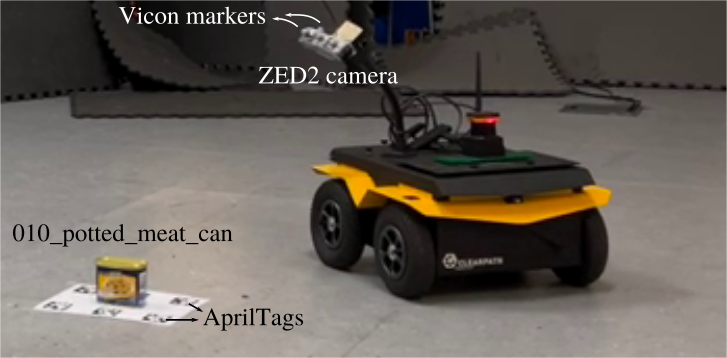}
    \captionsetup{justification=centering}
    \caption{The real robot experiment}
    \label{jackal}
\end{figure}

As illustrated in Fig.~\ref{jackal}, 
we control a Jackal robot \cite{jackal} to circle around two target objects: 003\_cracker\_box and 010\_potted\_meat\_can,
and collect stereo RGB images from a ZED2 camera \cite{ZED}.
For each object, two $\sim$4 min. long sequences are recorded, 
one for self-training and the other for testing.
We obtain the ground truth camera trajectory from a Vicon MoCap system \cite{Vicon}
and the ground truth object poses from AprilTag detections \cite{april}.
The camera odometry is computed with the SVO2 stereo module \cite{forster2014svo}.
We infer the object poses from the left camera RGB images using the same initial estimator as in Sec. IV.A.

\begin{figure}
    \centering
    \begin{subfigure}{0.45\linewidth}
        \centering
        \includegraphics[trim=30 20 30 0, clip, width=\linewidth]{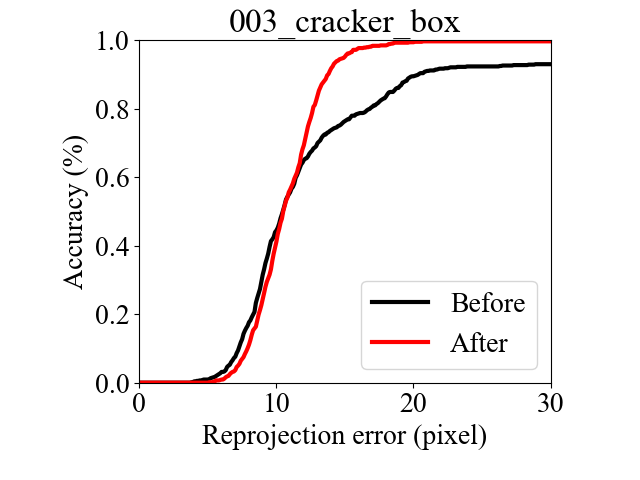}
    \end{subfigure}%
    \begin{subfigure}{0.45\linewidth}
        \centering
        \includegraphics[trim=30 20 30 0, clip, width=\linewidth]{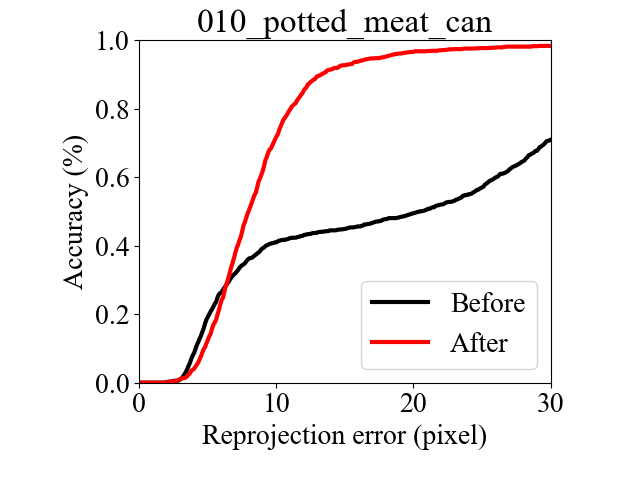}
    \end{subfigure}%
    \caption{
        \textbf{Quantitative evaluation} of our method on the test sets of the real robot experiment:
        Reprojection error accuracy-threshold curves before and after self-training.
        (Image size: 376$\times$672.)
    }
    \label{jackal_curve}
\end{figure}

Similar to the YCB-v experiment, 
we solve PGOs and pseudo-label the left camera images with the \textit{Hybrid} method.
For the two objects, 648 (out of 3120) and 950 (out of 2657) images are pseudo-labeled, in which 1.5\% and 2.8\% are hard examples.
After fine-tuning, we evaluate the estimator model on both self-collected test sequences and the YCB-v test sets.
On the self-collected test sequences, we adopt the reprojection error for the object 3D bounding box as the evaluation metric.
The accuracy-threshold curves are presented in Fig~\ref{jackal_curve}.
The AUCs for the curves in both tests are reported in Tab.~\ref{jackal_eval}.

We observe that the self-trained estimator makes considerably more (+17.75\% on average) accurate predictions in the same experimental environment.
More interestingly, in another unseen real-world domain, YCB-v test scenes, the self-trained model also shows enhanced prediction accuracy (+4.8 on average). 
We infer that this is because the domain gap between real environments is less than the sim2real gap.

\begin{table}[htb!]
    \centering

    \caption{
        \textbf{Quantitative evaluation} of our self-training method on the real robot experiment:
        the area under curve (AUC) values (\%) for the accuracy-threshold curves on the self-collected and the YCB-v test sets before and after self-training.
    }
    \begin{tabular}{c | c c }
        \hline
         \multicolumn{3}{c}{Self-collected test sequences} \\
        \hline
        Reproj error AUC & 003\_cracker\_box & 010\_potted\_meat\_can \\
        \hline
        Before          &     58.0          &    40.6                \\
        After           &     \textbf{63.9} &    \textbf{70.2}       \\
        \hline
        \hline
        \multicolumn{3}{c}{YCB-v testsets} \\
        \hline
        ADD AUC    & 003\_cracker\_box & 010\_potted\_meat\_can \\ 
        \hline
        Before    &  8.0          & 42.3 \\
        After & \textbf{15.2} & \textbf{44.7} \\
        \hline
    \end{tabular}
    \label{jackal_eval}
\end{table}

Similar to the YCB-v experiment, 
we further evaluate the SLAM performance gain after self-training.
We solve the PGOs on the self-collected test sequences and the YCB-v test set 0059, 
with the initial and the self-trained models' predictions.
As reported in Tab.~\ref{jackal_pgo}, the trajectory estimation errors are reduced in the same test environment and in another unseen YCB-v test domain.


\begin{table}[htb!]
    \centering
    \scriptsize
    \caption{
        \textbf{Object SLAM accuracy} on the self-collected test sequences and the YCB-v test set 0059 \textbf{before and after self-training}: absolute trajectory errors (ATE).
    }
    \begin{tabular}{c | c c c}
    \hline
    ATE (m) & 003\_cracker\_box test & 010\_potted\_meat\_can test & YCB-v 0059 \\
    \hline
    Before & 0.062          & 0.253          & 0.618\\
    After  & \textbf{0.046} & \textbf{0.072} & \textbf{0.576}\\
    \hline
    \end{tabular}
    \label{jackal_pgo}
\end{table}


\section{CONCLUSIONS}

We present a SLAM-aided self-training method for domain adaptation of 6D object pose estimators.
We exploit robust PGO results to generate multi-view consistent pseudo labels on robot-collected images and fine-tune the estimator models.
We propose an easy-to-implement robust PGO method, that can automatically tune the pose prediction covariances component-wise.
We evaluate our method on the YCB-v dataset and with real robot experiments.
The method can mine high-quality pseudo-labeled data from noisy and outlier-corrupted pose predictions.
After self-training, the pose estimators (initially trained on synthetic data) show considerably more accurate and consistent performance in the test domain, which also boosts the object SLAM accuracy.




\section*{APPENDIX}

\subsection{Proof for monotonic improvements in the joint loss}
We prove that our automatic covariance tuning method can monotonically reduce the joint loss $\Ljoint(\Variables, \Sigma_k)$ as defined in \eqref{joint}.
At iteration $i$, we have the variable assignments $\Variables^{(i-1)}$ and the noise covariances $\Sigma_k^{(i-1)}$ from iteration $i-1$. 
Solving the PGO, the LM algorithm ensures $\mathcal{L}(\Variables^{(i)}) \geq \mathcal{L}(\Variables^{(i-1)})$.
Since the extra regularization term in $\Ljoint$ is not a function of $\Variables$, 
we can have:
\begin{equation}
    \Ljoint(\Variables^{(i)}, \Sigma_k^{(i-1)}) \leq \Ljoint(\Variables^{(i-1)}, \Sigma_k^{(i-1)})
\end{equation}
We have also shown after \eqref{extremum} that $\Sigma_k = \text{diag}(|e_k|/\sqrt{\lambda})$ is a global minimizer for $\Ljoint$.
Thus, we have:
\begin{equation}
    \Ljoint(\Variables^{(i)}, \Sigma_k^{(i)}) \leq \Ljoint(\Variables^{(i)}, \Sigma_k^{(i-1)})
\end{equation}
Combining the two inequalities, we obtain:
\begin{equation}
    \Ljoint(\Variables^{(i)}, \Sigma_k^{(i)}) \leq \Ljoint(\Variables^{(i-1)}, \Sigma_k^{(i-1)})
\end{equation}
With this being valid for all iterations, we can have the chain of inequalities:
\begin{equation}
    \Ljoint^{(0)} \geq \Ljoint^{(1)} \geq \cdots \geq \Ljoint^{(N)}
\end{equation}
which completes the proof.

In our implementation, we eliminate outliers' influence by setting their $\Sigma_k$ to a large value (Alg.~\ref{altmin} line~\ref{explode}), 
which may violate this monotonic improvement property.
We choose to freeze the losses of outliers in $\Ljoint$ till the end of optimization to resolve the problem.

\subsection{Reduction to L1 robust M-estimator}
We show that our automatic covariance tuning method, as the noise models $\Sigma_k$ are assumed to be \textit{isotropic}, 
is equivalent to using the L1 robust M-estimator.
With the isotropic noises, i.e. $\Sigma_k = \sigma_k\mathcal{I}$, the joint loss in \eqref{joint} reduces to:
\begin{equation}
    \mathcal{L}_{\text{joint}} = \sum_{k} \frac{1}{\sigma_k^2}\|e_k\|_2^{2} + 6\lambda\sum_{k} \sigma_k^2 + \sum_{t} \|e_{t}\|_{\Sigma_{t}}^{2}
\end{equation}
Evaluating $\partial \Ljoint/\partial \sigma^2_k = 0$ yields the new update rule:
\begin{equation}\label{newupdate}
    \Sigma_k^{(i)} = \frac{1}{\sqrt{6\lambda}}|e^{(i)}_k|\mathcal{I}
\end{equation}

On the other hand, 
we can apply the L1 robust loss for the object pose measurement factors $f_k\in\Factors_Z$ to minimize the PGO loss \eqref{pgoloss}.
The robust PGO cost can be expressed as:
\begin{equation}\label{robustloss}
    \mathcal{L}_{\text{robust}} = \sum_{k} \rho(\|e_k\|_{\Sigma_k}) + \sum_{t} \|e_{t}\|_{\Sigma_{t}}^{2}
\end{equation}
where $\rho(\cdot) = |\cdot|$ is the L1 robust kernel.
The robust cost is typically minimized with the IRLS method,
by matching the gradients of \eqref{robustloss} locally with a sequence of weighted least squares problems.
The local least squares formulation at iteration $i$ can be expressed as:
\begin{equation}\label{irls}
    \mathcal{L}_{\text{IRLS}} = \sum_{k} w(\|e^{(i)}_k\|_{\Sigma_k})\|e_k\|_{\Sigma_k}^2 + \sum_{t} \|e_{t}\|_{\Sigma_{t}}^{2}
\end{equation}
where $w(\cdot) = 1/|\cdot|$ is the weight function.
Under the isotropic noise assumption, i.e. $\Sigma_k = \sigma_k\mathcal{I}$,
we can absorb the weight function into the covariance matrix and rewrite \eqref{irls} as:
\begin{equation}
    \mathcal{L}_{\text{IRLS}} = \sum_{k} \|e_k\|_{\Sigma_k^{(i)}}^2 + \sum_{t} \|e_{t}\|_{\Sigma_{t}}^{2}
\end{equation}
where the covariance matrix is de facto re-scaled iteratively by:
\begin{equation}\label{l1update}
    \Sigma_k^{(i)} = \sigma_k|e^{(i)}_k|\mathcal{I}
\end{equation}
Matching \eqref{newupdate} with \eqref{l1update}, we can see as $\frac{1}{\sqrt{6\lambda}} = \sigma_k$, 
the two methods are in theory equivalent\footnote{$\sigma_k$ = const. in the context of robust M-estimation.}.

\section*{ACKNOWLEDGMENT}
The authors acknowledge Jonathan Tremblay and other NVIDIA developers for providing consultation on training DOPE networks and generating synthetic data.
The authors also acknowledge the MIT SuperCloud and Lincoln Laboratory Supercomputing Center for HPC resources that have contributed to the results reported within this paper.


\bibliographystyle{IEEEtran}
\bibliography{root}

\end{document}